# Interpretable Machine Learning for Life Expectancy Prediction: A Comparative Study of Linear Regression, Decision Tree, and Random Forest

Roman Dolgopolyi, Ioanna Amaslidou, Agrippina Margaritou

{r.dolgopolyi, i.amaslidou, a.margaritou}@acg.edu

**Abstract.** Life expectancy is a fundamental indicator of population health and socio-economic well-being, yet accurately forecasting it remains challenging due to the interplay of demographic, environmental, and healthcare factors. This study evaluates three machine learning models—Linear Regression (LR), Regression Decision Tree (RDT), and Random Forest (RF), using a real-world dataset drawn from World Health Organization (WHO) and United Nations (UN) sources. After extensive preprocessing to address missing values and inconsistencies, each model's performance was assessed with $R^2$, Mean Absolute Error (MAE), and Root Mean Squared Error (RMSE). Results show that RF achieves the highest predictive accuracy ($R^2 = 0.9423$), significantly outperforming LR and RDT. Interpretability was prioritized through p-values for LR and feature-importance metrics for the tree-based models, revealing immunization rates (diphtheria, measles) and demographic attributes (HIV/AIDS, adult mortality) as critical drivers of life-expectancy predictions. These insights underscore the synergy between ensemble methods and transparency in addressing public-health challenges. Future research should explore advanced imputation strategies, alternative algorithms (e.g., neural networks), and updated data to further refine predictive accuracy and support evidence-based policymaking in global health contexts.

# 1    Introduction

Life expectancy is one of the most vital indicators of public health, economic development, and overall societal well-being. Over the last century, improvements in healthcare, sanitation, education, and living standards have contributed to a global rise in life expectancy. However, significant disparities remain between regions [1]. These discrepancies reflect underlying inequalities in access to healthcare, resources, and infrastructure. At the same time, emerging global challenges—such as pandemics, climate change, and rising rates of non-communicable diseases—continue to threaten gains in longevity. Understanding the underlying determinants of life expectancy is therefore essential not only for identifying vulnerable populations but also for designing targeted interventions and informing public policy [2].

Accurate life expectancy prediction plays a critical role in health planning and resource allocation. Policymakers, global health organizations, and healthcare providers rely on these forecasts to shape public health strategies, direct funding, and prioritize preventive measures [3]. Traditionally, life expectancy has been estimated using demographic models and life tables that rely heavily on historical mortality data [4]. While useful, these approaches often fall short in capturing the complex, non-linear interactions among health, environmental, and socioeconomic factors—especially in the face of fast-changing global conditions.

To address these limitations, the present study leverages machine learning (ML) techniques to analyze a real-world, multidimensional dataset compiled from the World Health Organization (WHO) and United Nations (UN) [5, 6]. This dataset encompasses a diverse set of features, including vaccination coverage, geographic and demographic data, mortality indicators, and socioeconomic variables across multiple countries and time periods. To model life expectancy based on these variables, three predictive algorithms of increasing complexity were developed and evaluated: Linear Regression (LR), Regression Decision Tree (RDT), and Random Forest (RF). Beyond identifying the most accurate and computationally efficient model, this study also emphasizes model interpretability by exploring the internal decision-making mechanisms of each algorithm. This focus aims to mitigate the "black-box" nature of ML techniques and enhance the transparency and reliability of their application in such critical areas as public health, where flaws in algorithms logic may lead to dangerous consequences [7].

# 2    Literature Review

Machine learning (ML) methods have been increasingly applied to life expectancy forecasting, using diverse inputs such as socioeconomic metrics, health indicators, and demographic variables to capture the complexity of global health outcomes [8]. Researchers have investigated a wide array of approaches, ranging from simpler linear regressions and single decision trees to more sophisticated models like random forests, gradient boosting and neural networks [9, 10, 11]. Although more advanced models typically exhibit enhanced predictive accuracy, studies consistently emphasize the importance of interpretability—particularly in sensitive domains like public health, where transparent decision-making is vital [12]. Numerous strategies have been proposed to tackle this challenge, including visualizing decision boundaries, extracting rule-based representations, and visualizing individuals layers of neural networks [13, 14, 15]. However, comparatively few works explore these interpretability strategies in depth for life expectancy models, leaving many policy-focused applications hesitant to rely on "black-box" predictions [16]. Consequently, there is growing interest in the systematic evaluation of both conventional and more sophisticated algorithms alongside interpretability techniques—especially for contexts where insights must be communicated to non-technical stakeholders [17]. In light of this gap, the present study compares models of varying complexity while simultaneously employing p-values, feature importance, and visual demonstration of the exact algorithm structure to ensure that high-accuracy models do not come at the expense of clarity and accountability for real-world health interventions.

## 3 Materials & Methods

### 3.1 Data Description

The dataset has been collected and merged from the WHO and UN repositories [5, 6]. It contains health and economic data for 193 countries, with a main focus on the relationship between life expectancy and health for the years 2000 – 2015. The merged version of the dataset can be accessed through Kaggle [18]. It consists of 22 columns and 2,938 rows. The features, as they appear in the file, are described in the following table (Table. 1).

**Table 1.** Feature description

| No. | Feature | Description |
| --- | --- | --- |
| 1 | Country | Name of the country (categorical) |
| 2 | Year | Year of data collection |
| 3 | Status | Development status of the country (Developed/Developing) |
| 4 | Life expectancy | Average number of years a person is expected to live |
| 5 | Adult Mortality | Probability of dying between age 15 and 60 per 1,000 population |
| 6 | Infant deaths | Number of deaths of children under 1 year per 1,000 live births |
| 7 | Alcohol | Alcohol consumption per capita (in litres) |
| 8 | Percentage expenditure | Expenditure on health as a percentage of GDP |
| 9 | Hepatitis B | Percentage of 1-year-old children immunized against Hepatitis B |
| 10 | Measles | Number of reported cases of Measles per 1,000 population |
| 11 | BMI | Average Body Mass Index of the population |
| 12 | Under-five deaths | Number of deaths of children under 5 years per 1,000 live births |
| 13 | Polio | Percentage of 1-year-old children immunized against Polio |

| 14 | Total expenditure | Total health expenditure as a percentage of GDP |
| 15 | Diphtheria | Percentage of 1-year-old children immunized against Diphtheria |
| 16 | HIV/AIDS | Deaths per 1,000 live births due to HIV/AIDS |
| 17 | GDP | Gross Domestic Product per capita (in USD) |
| 18 | Population | Total population of the country |
| 19 | Thinness (1–19 years) | Prevalence of thinness among children aged 10–19 |
| 20 | Thinness (5–9 years) | Prevalence of thinness among children aged 5–9 |
| 21 | Income composition of resources | Index combining income distribution and access to resources |
| 22 | Schooling | Average number of years of schooling received by people aged 25 or older |

### 3.2 Data Preprocessing

**Data loading**

. Data was loaded into a pandas DataFrame and all feature names were manually edited, since initially there were written with inconsistent syntax [19]. Renamed version of the features could be assessed in the following table (Table. 2).

Table 2. Feature renaming

| No. | Original | Renamed |
| --- | --- | --- |
| 1 | Country | country |
| 2 | Year | year |
| 3 | Status | status |
| 4 | Life Expectancy | life_expectancy |
| 5 | Adult Mortality | adult_mortality |
| 6 | Infant Death | Infant_deaths |
| 7 | Alcohol | alcohol |
| 8 | percentage expenditure | percent_expenditure |
| 9 | Hepatitis B | hepatitis_b |
| 10 | Measles | measles |
| 11 | BMI | bmi |
| 12 | under-five deaths | under_five_deaths |
| 13 | Polio | polio |
| 14 | Total expenditure | tot_expenditure |
| 15 | Diphtheria | diphtheria |
| 16 | HIV/AIDS | hiv_aids |
| 17 | GDP | gdp |
| 18 | Population | population |

| 19 | Thinness 1-19 years | thinness_1to19years |
|---|---|---|
| 20 | Thinness 5-9 years | thinness_5to9years |
| 21 | Income composition of resources | income_composition_of_resources |
| 22 | Schooling | school_years |

**Handling of missing values**
. The absence of missing values is essential to ensure the integrity and consistency of data analysis. Therefore, features containing more than 5% of missing values were excluded from the dataset. The excluded features were *alcohol, hepatitis_b, total expenditure, gdp, population, income_composition_of_resources, school_years*. Following this, all samples (rows) containing any remaining missing values were removed. This step resulted in the elimination of fewer than 2% of the total samples. After these preprocessing steps, the dataset was transformed into the size of 2,888 samples and 15 features, while contained no missing values.

**Handling of inconsistencies**
. To ensure the validity of the dataset, logical inconsistencies of the data were addressed through the application of upper-bound thresholds and internal consistency checks. For instance, values such as *life_expectancy* and *bmi* were limited to a maximum of 100, while features like *infant_deaths* and *under_five_deaths* were compared to ensure reasonable relationships between them. Such addressed inconsistencies may occur in large-scale datasets compiled by organizations like the WHO and UN due to variations in national reporting standards, manual data entry errors, or limitations in data collection processes [20]. After this step, the dataset size became—941 samples and 15 features, conforming to all necessary constraints for meaningful analysis.

**Mapping categorical features to numerical.**
. Next, mapping of the categorical features to numerical was performed. Firstly, the names of countries were replaced with their corresponding latitude and longitude values that were saved as separate features—*longitude, longitude*. These values were taken from the Google Earth's repository [21]. The remaining categorical feature *status* was transformed by applying the get_dummies function of the pandas library, resulting in two distinct features – *status_developing, and status_developed* [19].

### 3.3 Data Visualization

**Descriptive Statistics of Features**
. Additionally, distribution statistics of each feature were plotted for provision of manual and visual analyses. The histograms of features that provided the most insight was reported (Fig. 1 – 4).

6      R. Dolgopolyi et al.6      R. Dolgopolyi et al.

**Fig. 1.** Histogram of year

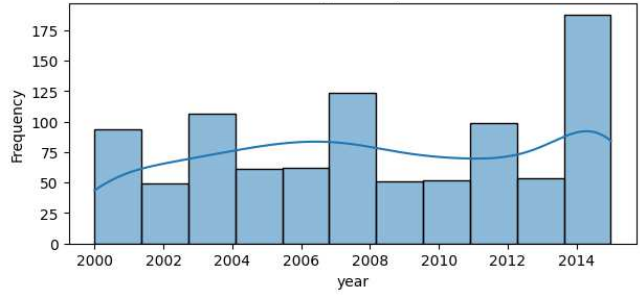

**Fig. 2.** Histogram of ifant deaths

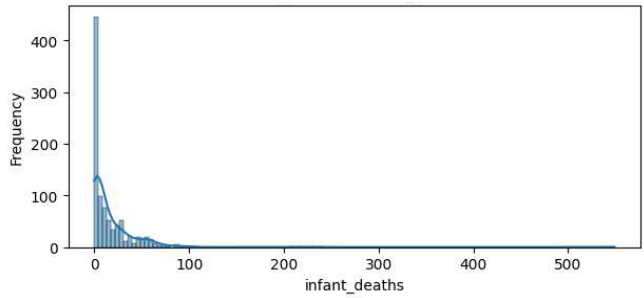

**Fig. 3.** Histogram of polio

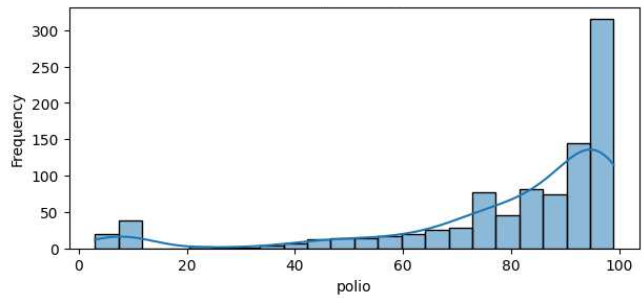

**Fig. 4.** Histogram of diphtheria

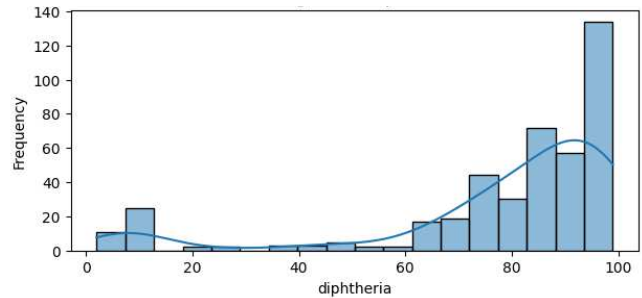

From the first histogram (Fig. 1) we may conclude that the amount of gathered socioeconomic and healthcare-related information was increasing globally from 2000 to 2015 year. The second histogram (Fig. 2) demonstrates that most of the countries over the same time period had relatively low rate of *infant_deaths,* however some significant outliers existed, which indicate uneven distribution to healthcare across the globe . The third histogram (Fig. 3) that describes the *polio* and the fourth (Fig. 4) that describes the *diphtheria* follow the identical distributions, which may suggest similar patterns across different vaccination programs during that time period.

**Descriptive Statistics of Target Variable**
. Distribution of the target variable of *life_expectancy* was plotted as well to provide an ability for visual representation and further analysis.

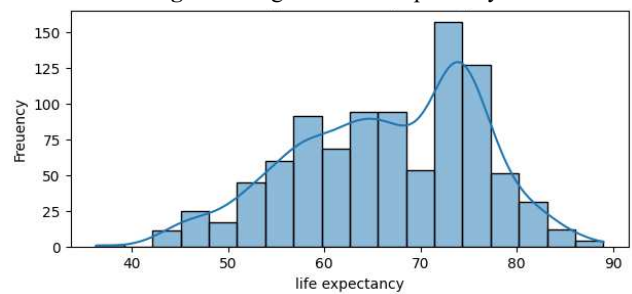

**Fig. 5.** Histogram of life expectancy

The histogram of life expectancy (Fig. 5) reveals a moderately right-skewed distribution, where most countries report life expectancy between 60 and 75 years, with a noticeable peak around 70–75 years. However, the presence of a long tail on the left suggests that there was still substantial number of countries that experienced low life expectancy during the observed time period from 2000 to 2015 year. This uneven distribution reflects global health inequalities of that time, where access to healthcare, sanitation, nutrition, and public health infrastructure remained highly variable across nations.

**Clustering and PCA analysis**
. Prior to applying clustering algorithms and Principal Component Analysis (PCA) for dimensionality reduction and data visualization, data scaling is necessary to ensure consistency in features contribution. To identify the most appropriate scaling method, three distinct scalers were evaluated: Standard Scaler, Min-Max Scaler, and Max Abs Scaler. For each dataset with different scaling methods, clustering was conducted using both K-Means and Hierarchical clustering techniques, and the corresponding silhouette scores were calculated to assess clustering performance. Ultimately, only the clustering result with the highest silhouette score was retained and considered for further analysis.

Following the implementation of these methods, the optimal number of clusters was determined to be three, corresponding to the highest silhouette score of 0.59. To illustrate the relationship between the number of clusters and clustering quality, a silhouette score plot was generated (Fig. 6). This visualization reveals two distinct phases—an initial sharp increase in silhouette score from 1 to 3 clusters, followed by a gradual decline as the number of clusters increases from 4 to 50.

**Fig. 6.** Visualization of the silhouette scores for different number of clusters (k value).

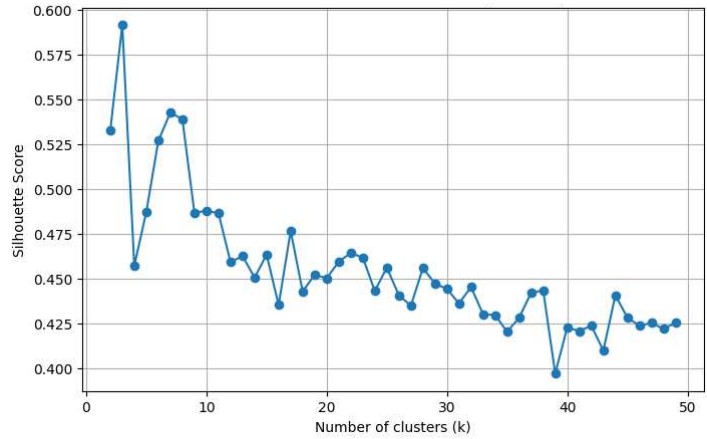

To further interpret and validate the clustering structure identified by the K-Means algorithm, a two-dimensional visualization was created using Principal Component Analysis (PCA). The resulting scatter plot (Fig. 7) presents the data projected onto the first two principal components, allowing for an intuitive representation of the clusters.

**Fig. 7.** Visualization of K-Means clusters using PCA.

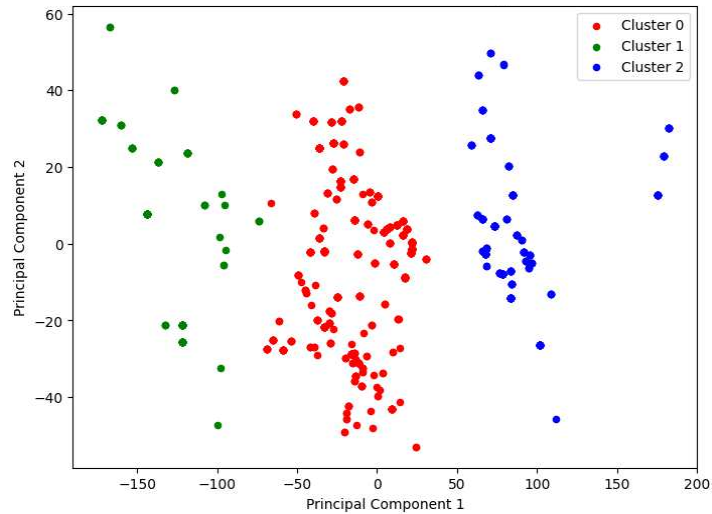

Low degree of overlap between the clusters and the distinct spatial separation observed in the PCA space provided visual confirmation that the K-Means algorithm with 3 clusters was indeed the optimal choice for the clustering process. The insight that the Max Abs scaling method transformed the data in the most suitable form for classification tasks was noted and it was decided to use this particular scaling method for further classification and prediction processes of the research.

**Feature Associations**

. Lastly, feature associations were assessed both individually and in relation to the target variable by calculating Pearson correlation coefficients. These correlations provided insights into the linear dependencies among numerical features, as well as their potential influence on the prediction target. To enhance interpretability, the results were visualized using heatmaps (Fig. 8 – 9), offering a clear and intuitive overview of the strength and direction of feature relationships. Furthermore, to evaluate associations between categorical and numerical variables, Analysis of Variance (ANOVA) was employed, as it quantifies the extent to which variation in a numerical feature can be explained by group differences in a categorical feature.

**Fig. 8.** The feature associations matrix visualized as a heatmap.

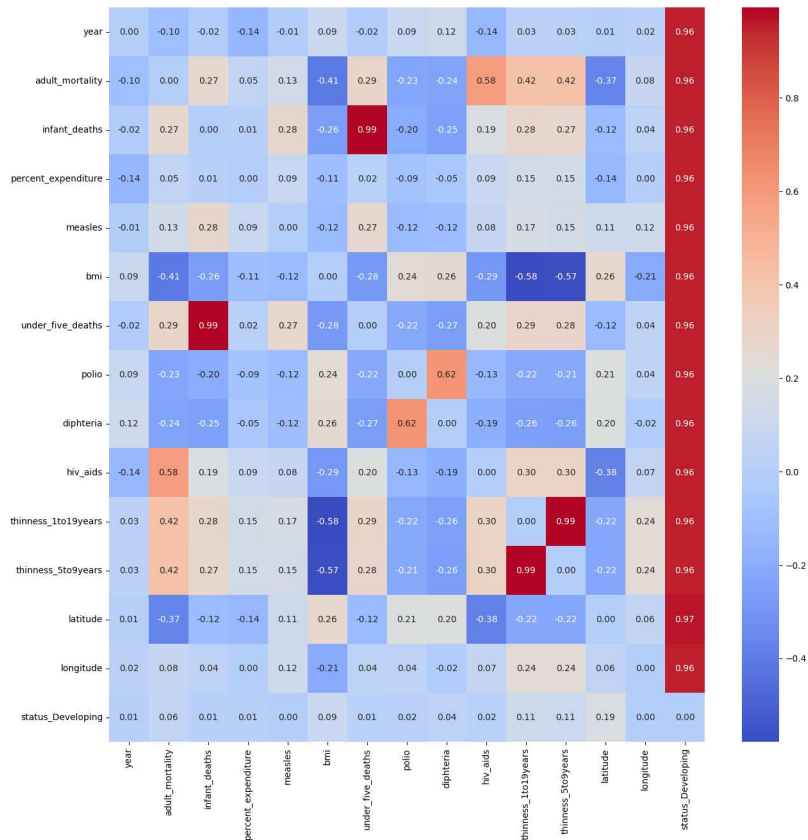

Thus, the feature associations matrix (Fig. 8) displays the Pearson correlation matrix among all numerical features in the dataset. Several strong positive and negative correlations are immediately apparent. The highest positive correlations are observed between *infant_deaths* and *under-five deaths* (r = 0.99), as well as between *thinness_1to19years* and *thinness_5to9years* (r = 0.99). These high values are expected, given the conceptual similarity and overlap in these health indicators, suggesting the validity of the previous preprocessing steps.

On the other hand, significant negative correlations include those between *bmi* and both thinness features (r = -0.58), indicating an inverse relationship where higher *bmi* values are typically associated with lower levels of thinness. Another notable negative correlation is found between *adult_mortality* and *diphtheria* vaccination coverage (r = -0.47), which suggests that increased vaccination rates may contribute to lower mortality among adults. The remaining coefficients range from mild to moderate in magnitude, highlighting more complex interdependencies among features that warrant further multivariate analysis.

**Fig. 9.** The feature associations to the target variable matrix visualized as a heatmap.

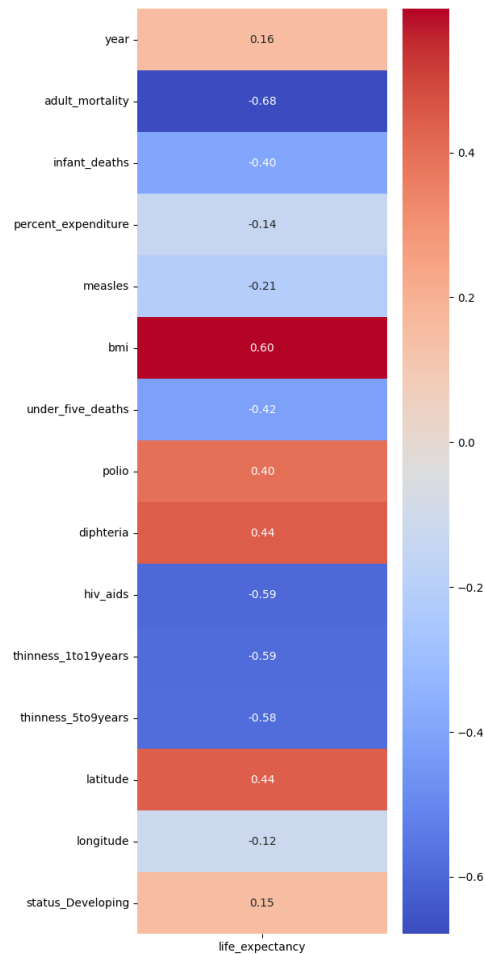

Lastly, the feature associations to the target variable (Fig. 9) shows various strong and weak correlations. Features that seem to decrease life expectancy include

*adult_mortality* (-0.68), *hiv_aids* (-0.59), *thinness_1_to_19years* (-0.59) and *thinness_5to years* (-0.58), while features like *bmi* (0.60), *diphtheria* (0.44), *polio* (0.40), *latitude* (0.44) exhibit positive correlation as they increase life expectancy.

### 3.4 Modelling

**Linear Regression**
. To fit the LR model, the dataset was divided into training and testing subsets using a 70:30 split. A validation dataset was not created, nor were cross-validation techniques applied, as the LR model does not require hyperparameter tuning. Additionally, a constant term (intercept), necessary for the LR formulation, was manually added to the feature set prior to model training to ensure proper fitting.

The model was implemented using the LinearRegression class from the Scikit-learn library [22]. Performance was evaluated on both the training and testing datasets using standard regression metrics, including the coefficient of determination ($R^2$), Mean Squared Error (MSE), Mean Absolute Error (MAE), and Root Mean Squared Error (RMSE). To further support the quantitative analysis, "Actual vs. Predicted" scatter plot (Fig. 10) were generated for the testing dataset, providing a visual assessment of the model's predictive accuracy and alignment with observed values. In addition, P-Values were computed (Fig. 11) to evaluate the significance of each individual feature for target variable prediction and to address model interpretability. Both training and inference time were recorded to assess computational complexity of the model.

**Regression Decision Tree**
. To implement the RDT model, the dataset was also divided into training and testing subsets using a 70:30 split. The model development involved hyperparameter tuning using a grid search strategy. The GridSearchCV class from the Scikit-learn library, configured with 5-fold cross-validation, was used to identify the optimal set of hyperparameters, including maximum tree depth, minimum number of samples per leaf, cost-complexity pruning parameter, and the splitting criterion [22]. The grid search was set to optimize the $R^2$ score as the primary scoring metric.

Once the best-performing configuration was identified, the model was retrained on the full training set. Performance on the testing dataset was evaluated using $R^2$, MAE, MSE, and RMSE (Table. 3). "Actual vs. Predicted" scatter plot for the performance on the testing dataset was generated to visually assess the alignment between predicted and actual values (Fig. 12). Additionally, to enhance model interpretability, the trained decision tree was visualized using the plot_tree function from Scikit-learn (Fig. 13) [22]. Training and inference time were also measured to evaluate computational complexity (Table. 3).

**Random Forest**
. The RF model was developed using the same 70:30 train-test split as in the previous models. In this implementation, the best-performing hyperparameters from the RDT

model—such as maximum tree depth, minimum number of samples per leaf, cost-complexity pruning parameter, and the split criterion—were retained and applied to the RF model.

Further hyperparameter tuning was conducted specifically for the RF parameters using the GridSearchCV class with 5-fold cross-validation [22]. The grid search focused on optimizing the number of trees, the number of features considered at each split, and the use of bootstrap sampling. The $R^2$ metric was used to guide model selection.

After identifying the optimal configuration, the final model was retrained on the full training dataset and evaluated on the testing set using $R^2$, MAE, MSE, and RMSE (Table. 3). "Actual vs. Predicted" scatter plot was produced to visually inspect model performance on the testing dataset, and feature importance scores were extracted to highlight the relative influence of each input variable for provision of the model interpretability. Training and inference times were recorded as part of the evaluation of the model's computational complexity (Table. 3).

## 4  Results

The table below (Table. 3) summarizes the performance outcomes for the three predictive models—LR, RDT, and RF—evaluated on the testing dataset. The results are presented in terms of $R^2$, MAE, RMSE, as well as training and inference time, providing a comprehensive view of both predictive accuracy and computational complexity. As observed, the models exhibit distinct trade-offs between computational complexity and predictive performance.

**Table 3.** Evaluation Results of Predictive Models on Test Dataset

| Model | $R^2$ | MAE | RMSE | Training Time (s) | Inference Time (s) |
|---|---|---|---|---|---|
| LR | 0.7296 | 3.7398 | 4.8169 | 0.00235 | 0.00002 |
| DRT | 0.8979 | 2.0902 | 2.9602 | 11.4200 | 0.00005 |
| RF | 0.9423 | 1.4799 | 2.2245 | 32.6858 | 0.00358 |

The following sections provide a detailed discussion of each model's performance based on these metrics, highlighting the strengths and limitations of each approach in the context of life expectancy prediction, computational complexity, and model interpretability.

**Linear Regression**
. LR model, while offering the lowest computational cost among the three approaches, it demonstrated relatively modest predictive performance. On the testing dataset, it achieved an $R^2$ score of 0.7296, indicating that approximately 73% of the variance in life expectancy could be explained by the model. The corresponding MAE and RMSE values were 3.7398 and 4.8169, respectively, which suggest that the model's average and squared prediction errors were higher compared to the tree-based models. Despite this, the model's simplicity enabled extremely fast training and inference, with training completed in just 0.00235 seconds and per-sample inference averaging 0.00002 seconds. These results make the LR model attractive for applications where computational resources are highly constrained or rapid responses are prioritized over prediction precision.

Visual inspection of the model's predictions through the "Actual vs. Predicted" scatter plot (Fig. 10) reveals a general alignment along the identity line, though notable dispersion is observed, particularly in cases with lower or higher life expectancy values. This suggests that while the model captures the overall trend, it may struggle with capturing non-linear patterns in the data.

**Fig. 10.** Actual vs Predicted (Linear Regression – Test Set)

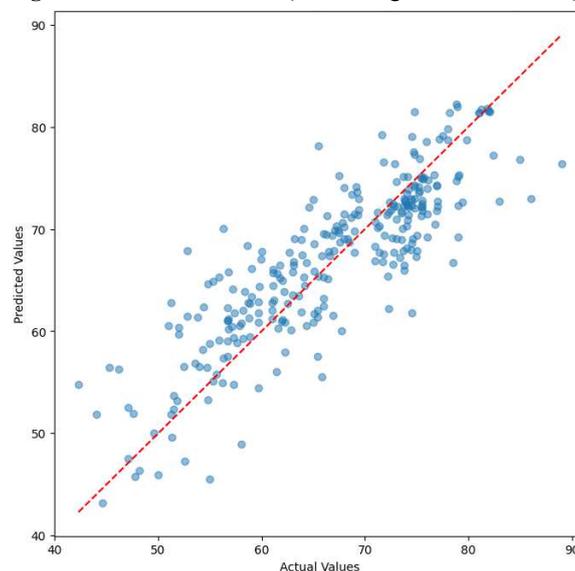

To further analyze the model's explanatory variables, p-values for each feature's regression coefficient were examined (Fig. 11). Features that exhibited statistically significant contributions ($p < 0.05$) were identified in significance order: *adult_mortality, hiv_aids, bmi, status_developing*, *diphtheria*, *polio, year, latitude, measles, under_five_deaths, infant_deaths*. All of these features indicated a strong relationship with the target variable. In contrast, features such as *thinness_1to19years, longitude, thinness_5to9years,* and *percent_expenditure* had high p-values, suggesting that they may not be significant predictors. This statistical insight helps identify which variables meaningfully contribute to *life_expectancy* predictions within the LR framework, although the model remains limited in its ability to capture non-linear patterns and feature interactions.

**Fig. 11.** P-values for Features in LR Model

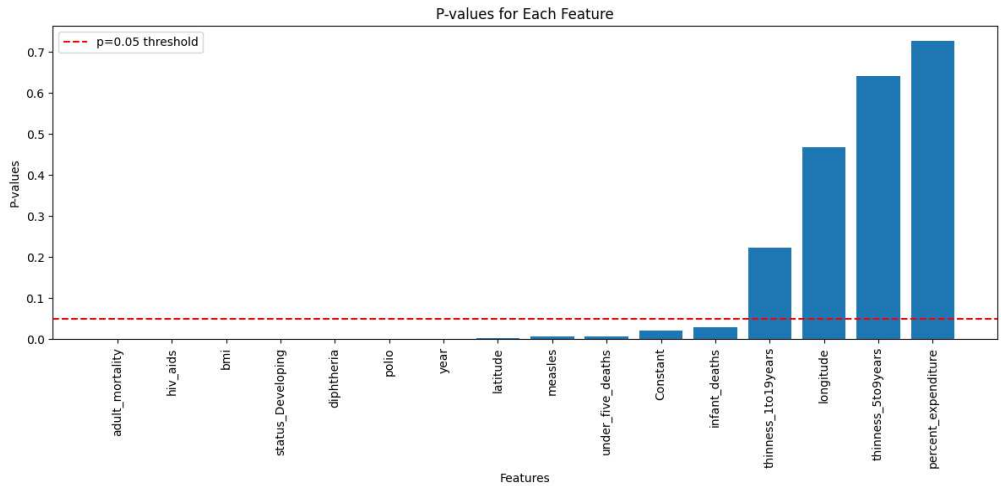

**Regression Decision Tree**

. The RDT model outperformed the LR model across all predictive accuracy metrics while maintaining moderate computational efficiency. On the test set, the RDT achieved an R² score of 0.8979, indicating a substantial improvement in the model's ability to explain variance in life expectancy outcomes. The corresponding MAE and RMSE values were 2.0902 and 2.9602, respectively, demonstrating significantly reduced average and squared prediction errors compared to the LR model. These improvements reflect the RDT's capacity to capture non-linear relationships and interactions between features that linear models inherently miss.

Hyperparameter tuning was conducted using a grid search with 5-fold cross-validation. The optimal configuration was achieved with a maximum tree depth of 15, a minimum of 5 samples required at a leaf node, a cost-complexity pruning parameter of 0.01, and the squared error criterion for measuring the quality of splits. This set of parameters allowed the model to maintain sufficient complexity to fit the training data while preventing overfitting. The total training time, including the hyperparameter tuning process, was 11.42 seconds, with the inference time of 0.00005 seconds per sample.

The "Actual vs. Predicted" plot (Fig. 12) reveals a tighter clustering of predictions around the identity line compared to LR, indicating higher predictive alignment.

**Fig. 12.** Actual vs Predicted (Regression Decision Tree – Test Set)

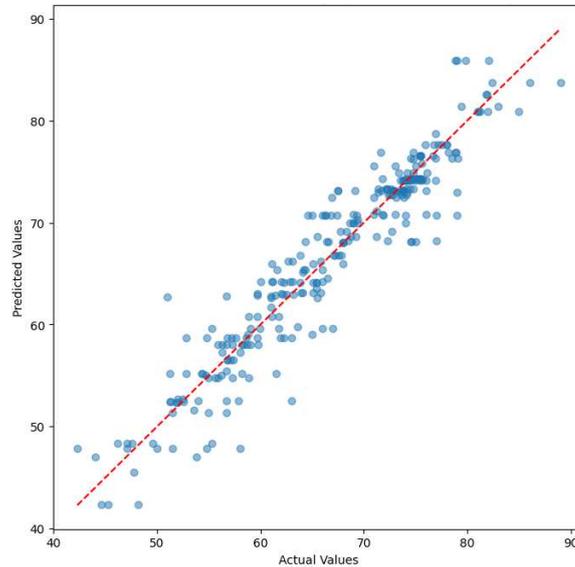

Additionally, a visualization of the tree structure (Fig. 13) provided a clear and interpretable breakdown of how the model uses feature thresholds to guide predictions. The tree prominently utilized features are *hiv_aids*, *adult_mortality*, *longitude*, *under_five_deaths, status_developing*.

**Fig. 13.** Regression Decision Tree – Structure of First 3 Levels

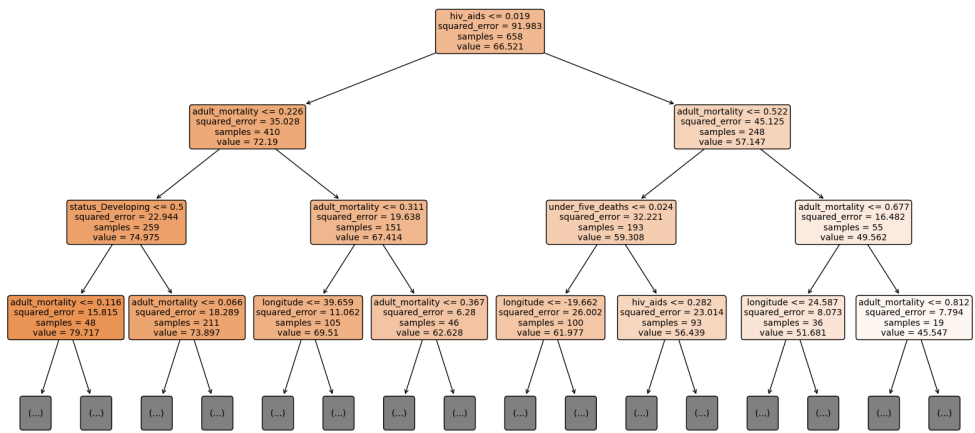

**Random Forest**

. The RF model demonstrated the highest predictive performance among the three approaches, achieving an R² score of 0.9423 on the test dataset. This indicates that the model was able to explain over 94% of the variance in life expectancy, outperforming both the Decision Tree and Linear Regression models. It also recorded the lowest MAE (1.4799) and RMSE (2.2245), confirming its superior accuracy in both average and squared prediction errors. These results highlight the model's ability to learn complex, non-linear relationships from the data more effectively than the simpler alternatives.

The model was initially configured using the best hyperparameters identified during DRT tuning, including a maximum tree depth of 15, a minimum of 5 samples per a tree leaf and a pruning threshold of 0.01, with squared error used to evaluate split quality. Further hyperparameter tuning was performed specifically for the RF, using a grid search with 5-fold cross-validation. The optimal configuration included 200 decision trees, no bootstrap sampling, the same squared error criterion for splits, and a requirement of at least 2 samples to create a split at each node. The total training time, including the tuning process, was 32.69 seconds—higher than for the other models—while prediction time per sample remained reasonably efficient at 0.00358 seconds.

The "Actual vs. Predicted" scatter plot (Fig. 14) illustrates a strong alignment between predicted and true values, with minimal dispersion, indicating excellent predictive reliability.

**Fig. 14.** Actual vs Predicted (Random Forest – Test Set)

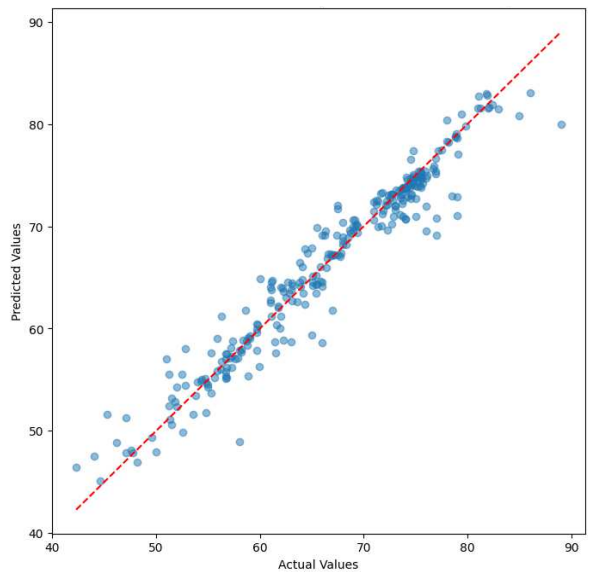

Feature importance analysis (Fig. 15) showed that variables such as *diphtheria*, *year*, and *measles* were the most influential in the model's predictions. In contrast, variables that were highly impactful in other models, such as *hiv_aids*, or *adult_mortality*, played a lesser role in this model's decision-making. This divergence highlights the RF tendency to prioritize patterns differently, depending on its structure and training. Despite its relatively greater computational demands, the RF model ultimately proved to be the most accurate and robust solution for predicting life expectancy.

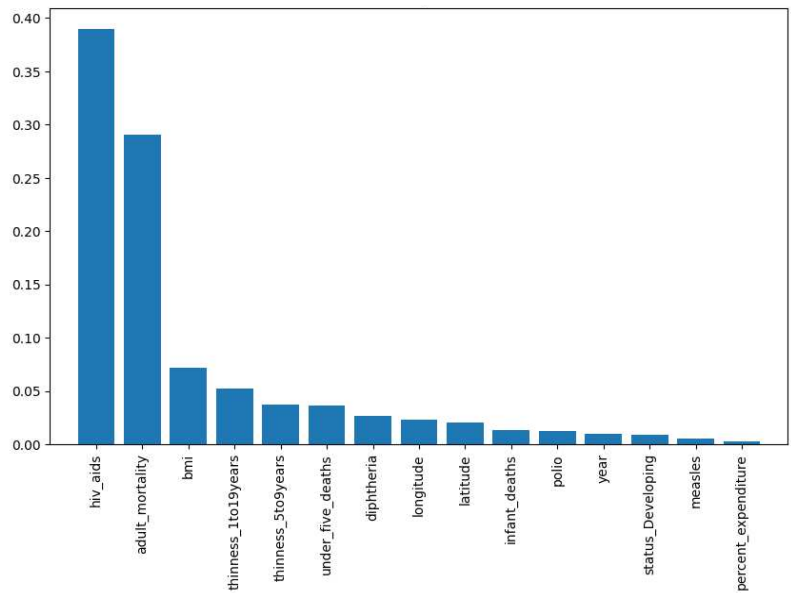

**Fig. 15.** Features importance – Random Forest

## 5 Conclusion

In this study, we compared three machine learning models of varying complexity—LR, RDT, and RF—to predict life expectancy using a multidimensional dataset sourced from the WHO and UN [5, 6]. Our findings indicate that while LR provides a computationally inexpensive solution, it underperforms when confronted with the dataset's inherent non-linearities. By contrast, the tree-based models exhibit superior predictive accuracy, with the RF ultimately achieving the highest $R^2$ of 0.9423 on the testing dataset. This result underscores the value of ensemble methods for handling complex relationships among health, socioeconomic, and demographic factors.

Importantly, this work prioritizes interpretability. Although more advanced models often yield stronger predictive outcomes, they may obscure the contribution of individual features—undermining stakeholder trust in high-stakes areas like public health. To address this concern, we employed p-values in LR, visual examinations of tree structure, and feature-importance metrics for RF. These techniques illuminated how certain variables—*diphtheria, year, measles, hiv_aids*—can substantially shape life-expectancy estimates. Understanding these determinants can guide resource allocation, inform policy interventions, and enhance the transparency of AI-driven decision-making.

## 6 Limitations

Despite offering valuable insights into life-expectancy prediction, this study faces several constraints. First, the dataset is derived from the WHO and UN for the years 2000–2015. Although it provides extensive geographic, demographic, and health-related features, these findings may not fully capture recent global shifts in healthcare or the impact of emerging challenges, such as novel pandemics and evolving socio-economic conditions [24]. The analysis should therefore be interpreted with caution when extrapolating to current contexts.

Second, the decision to exclude features with more than 5% missing data and to remove rows containing any missing values may limit the breadth of the remaining dataset. Potentially relevant variables (e.g., GDP, population) were excluded, probably obscuring additional factors that influence life expectancy. Imputation or alternative data-handling strategies might yield different results, particularly in resource-constrained regions where data collection can be uneven.

Third, while multiple —LR, RDT, and RF—were investigated, the choice was not exhaustive. Other advanced algorithms, including gradient boosting or neural networks, might provide further improvements in predictive accuracy or interpretability.

Future work should address these limitations by updating the dataset to reflect ongoing demographic and healthcare changes, integrating richer imputation strategies, and expanding the range of algorithms tested.